\pdfoutput=1

\documentclass[11pt]{article}
\usepackage{graphicx}
\usepackage{acl}

\usepackage{times}
\usepackage{latexsym}

\usepackage[T1]{fontenc}

\usepackage[utf8]{inputenc}

\usepackage{microtype}
\usepackage{appendix}
\usepackage{etoolbox}
\BeforeBeginEnvironment{appendices}{\clearpage}
\usepackage{booktabs}

\title{MockingBERT: A Method for Retroactively Adding Resilience to NLP Models\footnotemark[1]  }

\author{Jan Jezabek \\
  Hedgefrog Software LLC \\
  \texttt{jjezabek@hedgefrogsoft.com} \\\And
  Akash Singh \\
  Industries Cloud, Salesforce Inc. \\
  \texttt{singh.akash@salesforce.com} \\}

\begin{document}
\maketitle
\begin{abstract}
Protecting NLP models against misspellings whether accidental or adversarial has been the object of research interest for the past few years. Existing remediations have typically either compromised accuracy or required full model re-training with each new class of attacks. We propose a novel method of retroactively adding resilience to misspellings to transformer-based NLP models. This robustness
can be achieved without the need for re-training of the original NLP model and with only a minimal loss
of language understanding performance on inputs without misspellings. Additionally we propose a new efficient approximate method of generating adversarial misspellings, which significantly reduces the cost needed to evaluate a model's resilience to adversarial attacks.
\end{abstract}

\footnotetext[1]{MockingBERT is a reference to mockingbirds, a group of birds known for mimicking the sounds of other animals. The code and data used for this article is publicly available on GitHub, HuggingFace Hub and S3.}

\section{Introduction}

While artificial neural networks have been able to achieve human level performance on many real-world tasks, they sometimes fail in surprising ways. \cite{szegedy2013intriguing} showed that state of the art computer vision models can be fooled into misclassifying objects with only limited perturbations imperceptible to human viewers. Along similar lines, it was shown in \cite{pruthi-etal-2019-combating} that very constrained attacks can successfully trick classification algorithms into making incorrect predictions. In fact such attacks have been used for a long time, chiefly for evading spam classifiers while remaining legible to human readers.

Protecting against such misspellings, whether accidental or intentional, has been a focus of research in the NLP field for many years \citep{Lee2005SpamDU}. Recently, defenses suggested by \cite{pruthi-etal-2019-combating} and \cite{jones-etal-2020-robust} can partially remediate adversarial attacks by adding a pre-processing step, at the cost of a drop in classification performance. \cite{Liu_Zhang_Wang_Lin_Chen_2020} proposed replacing a fixed word embeddings with trained character-based ones and observed improved resilience to adversarial attacks.

In existing systems a tension exists between modularity and accuracy. \cite{pruthi-etal-2019-combating} and \cite{jones-etal-2020-robust} propose fully modular systems that are completely oblivious of the downstream language understanding model. This provides explainability (by providing a verbatim sequence of corrected tokens) but comes at a cost of reduced accuracy on unperturbed inputs. Additionally there is an added drawback of not being able to preserve potential ambiguity present in the input, making these systems `destructive'. Conversely, \cite{Liu_Zhang_Wang_Lin_Chen_2020} is able to represent ambiguous inputs, however at the cost of losing modularity.

Our central hypothesis is that original accuracy can be preserved while at the same time ensuring modularity. In particular we show that existing classifiers based on BERT and RoBERTa, two widely used pre-trained models, can be retroactively made resilient to perturbations even if only unperturbed data was used during the initial finetuning. This can be done by replacing the heuristic subword tokenizer and token embedding with a machine learned replacement which we call MockingBERT. The MockingBERT tokenizer and embedder learns to mimic a transformer model's tokenization and layer 0 embedding mechanism while providing resilience to input perturbations.




We evaluate the performance of such models when trained on both unperturbed and perturbed training sets to understand their suitability for data augmentation. We perform a comparative analysis with the methods proposed in \cite{jones-etal-2020-robust}, as well as with a regular finetuned BERT model trained with data augmentation.

We also propose and evaluate \textsc{WordScoreAttack}, an efficient and effective method for generating adversarial samples without the need for exhaustively considering all possible perturbations. \textsc{WordScoreAttack} works  by carefully choosing information bearing words in the input text. This provides a significantly faster alternative to the exhaustive method proposed by \cite{pruthi-etal-2019-combating} at the cost of perturbing a larger number of characters in the sentence. Crucially this method requires significantly fewer calls to the underlying NLP model, which more closely approximates real world scenarios that are likely to involve rate limiting and/or a limited query budget.
We manually verify that the outputs of this perturbation can still be classified correctly by a human reader with a high probability.



Our central findings are that the MockingBERT tokenizer and embedder model paired with a finetuned BERT or RoBERTa classifier achieves a high level of resilience to character-level adversarial perturbations when pre-trained on perturbed data. This results in a higher accuracy on perturbed inputs on multiple well known datasets than state of the art methods for combating adversarial misspellings such as the one described in \cite{jones-etal-2020-robust}. Crucially, the impact on accuracy for unperturbed inputs, while measurable, is significantly lower than for comparable methods for protecting against adversarial attacks.


Additionally, \textsc{WordScoreAttack} significantly reduces the number of model queries required to find adversarial inputs. This dramatically speeds up evaluation, while also making the attack more practical in real scenarios.

\section{Related Work}

Previous research has explored both adversarial attacks against NLP systems as well as possible defenses. This section gives a brief overview of related work and concepts.

\subsection{Adversarial Attacks}
\cite{pruthi-etal-2019-combating} proposes an attack that exhaustively searches for a modification of a single word with the intent of causing a wrong prediction. The allowed modifications are from a narrow set: Adding or deleting a single internal character, swapping two neighboring internal characters or replacing an internal character with one of its neighbors in the QWERTY keyboard layout. This choice of perturbations was based on linguistic research which suggests that modifications to internal characters in a word do not significantly hinder legibility for a human reader \citep{rawlinson1976significance}.

Another approach to adversarial attacks works by replacing individual characters with similarly looking symbols or letters from different alphabets. In this scenario the text remains easily understandable to the reader even in the presence of a large number of misspellings \citep{eger-etal-2019-text, sokolov2020visual}.

\subsection{Defenses Against Attacks}
In addition to the attack described previously, \cite{pruthi-etal-2019-combating} proposes a remediation in the form of a subcharacter recurrent neural network (ScRNN), which attempts to reverse any perturbations present in the input sentence. \cite{jones-etal-2020-robust} proposes a system named RobEn that clusters misspellings of vocabulary words with a bounded edit distance and maps them to the most frequent word in the cluster. This approach works even in the case where every word in the sentence is misspelled, at the cost of reduced accuracy on unperturbed inputs.

Both of these approaches are highly modular, i.e. they can be used with any language understanding model as a preprocessing step. In contrast \cite{Liu_Zhang_Wang_Lin_Chen_2020} proposes joinly training a character-based word embedder and the main NLP model. This allows for the representation of ambiguous misspellings and their handling in the NLP model.

A similar approach is also considered in \cite{el-boukkouri-etal-2020-characterbert} where a character-level word embedding is passed to a transformer model with a number of parameters similar to BERT. Even though this approach is not specifically aiming for resilience to misspellings, the authors observe such resilience as a side effect of their embedding procedure.

\subsection{Alternatives to Subword Tokenization}

Our research specifically targets NLP models based on the transformer architecture \citep{vaswani2017attention}, with a focus on models derived from BERT \citep{devlin-etal-2019-bert} and RoBERTa \citep{DBLP:journals/corr/abs-1907-11692}. Such models are pre-trained on large corpora of unlabeled data, and can then be adapted (in a process known as finetuning) to many NLP tasks. This substantially reduces the amount of data needed for each individual task.

An important aspect of understanding BERT is that it uses the WordPiece subword tokenization \citep{wu2016google} process. This means that while common words are typically mapped to a single token, uncommon or invented words can still be represented by a sequence of tokens without the need to resort to a catch-all token (typically denoted as UNK). During the operation of the transformer model, tokens corresponding to subwords are first represented as context-free vectors of numbers (that had been learned during training), with subsequent layers incrementally adding contextual information from other tokens using the self-attention mechanism.

There are known alternatives to subword embeddings that are similarly able to avoid emitting out-of-vocabulary tokens. One prominent example is ELMo \citep{peters-etal-2018-deep}, which uses character-level embeddings in addition to a word's surrounding context to come up with an embedding for a particular word in a sentence. Similar techniques have been successfully used with transformer based models \citep{el-boukkouri-etal-2020-characterbert,ma-etal-2020-charbert}.

The tokenization schemes described so far all rely on whitespace tokenization, which means they are likely to be susceptible to adversarial attacks that insert or remove whitespace. In contrast some recent transformer based models avoid using heuristic tokenizers altogether while still using a comparable number of parameters to BERT \cite{clark2022canine}, \cite{tay2021charformer}. This is achieved by using a convolution-like process to map the sequence of input characters to embeddings before passing them to the transformer blocks.

\section{Data}

For training the MockingBERT tokenizer and embedder we use the unlabeled BookCorpus dataset \citep{zhu2015aligning}, with minimal pre-processing to reverse the existing tokenization present in that dataset. The goal is to utilize textual data that is as close as possible to unprocessed text.

For evaluation purposes we use the Large Movie Review (IMDb) dataset \citep{maas-etal-2011-learning}, the Stanford Sentiment Treebank (SST) dataset \citep{socher-etal-2013-recursive} and the Large Yelp Review dataset \citep{NIPS2015_250cf8b5}. For the latter two datasets we use both the 2-class (binary) and 5-class variants. We subsequently refer to these datasets as IMDb, SST-2, SST-5, Yelp-2 and Yelp-5.

For each task we evaluate each model's accuracy on a randomly chosen subset of 500 sentences from each dataset's test set. The reason for the limited size of the test set is that adversarial attacks require potentially hundreds of model inferences for each sentence to find a successful perturbation. 



\section{Models}

\begin{figure*}[ht]
    \centering
    \makebox[\columnwidth]{\includegraphics[width=16.0cm]{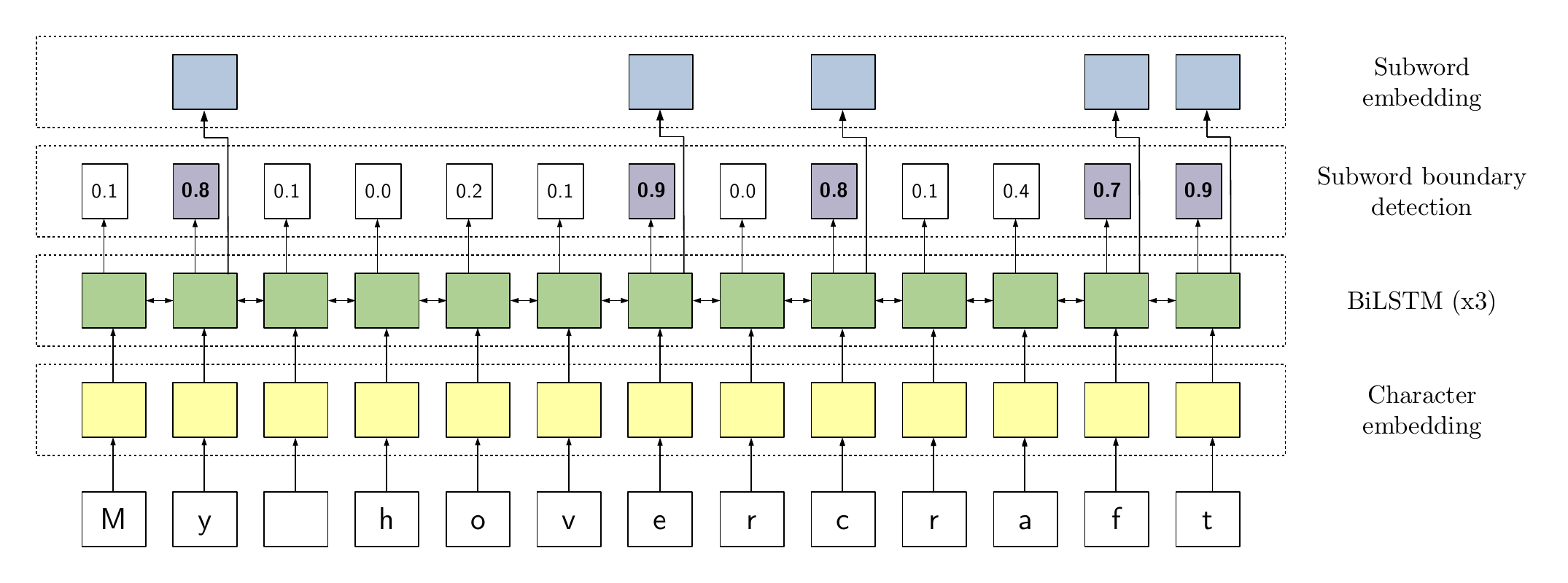}}
    \caption{Structure of the MockingBERT tokenizer and embedder. In this example the phrase `My hovercraft' is split into five tokens. An embeddings is only used if the corresponding subword boundary is detected with confidence of at least 0.5.}
    \label{fig:model_structure}
\end{figure*}


We propose a model, MockingBERT, that can be used in place of a transformer model's tokenizer and word embedding. The model consists of a character embedding layer that transforms each input character into a numeric vector of size 768. Characters are converted to lower case before embedding, but no other preprocessing is done (e.g. no special handling for whitespace or punctuation). The layer is followed by three stacked bidirectional LSTM layers, with a hidden size of 768. The final LSTM layer is connected to two parallel dense layers: The subword boundary detection layer and the subword embedding layer (Figure \ref{fig:model_structure}). The subword boundary detection layer uses a sigmoid activation function and has an output dimension of 1. The subword embedding layer uses hyperbolic tangent as its activation function and has an output dimension of 768, matching the subword embedding size of both BERT Base and RoBERTa Base.

The model is trained with two objectives. The first objective is a character sequence classification task which recognizes subword boundaries, i.e. the last character of a subword. This is used to break the input text into subwords. The second objective constructs the subword embedding for each subword.

The models are trained without supervision on sentences from BookCorpus \citep{zhu2015aligning}. During training an unperturbed sentence is processed by the transformer model's regular tokenizer and its context free word embedding (trained layer zero embeddings without the position and segmentation embeddings).  This allows us to determine the subword boundaries and subword embeddings. The sentence may subsequently be perturbed by adding or deleting a character, swapping two neighboring characters, inserting whitespace in the middle of long words, or removing whitespace between two words. The subword boundaries are updated accordingly. In the case of character deletion, special handling is present for characters that are subword boundaries. In this case the immediately preceding character is marked as a subword boundary. In the uncommon case when the preceding character is already a subword boundary, the embedding for the deleted character is simply discarded. In our opinion this situation is sufficiently rare that it does not warrant adding more complexity to the models.

When training, we can choose which context-free subword embedding MockingBERT will approximate:
\begin{itemize}
    \item We can target the embedding of the generic (that is not finetuned) BERT or RoBERTa model. An advantage of this is that the trained MockingBERT embedder is independent of the finetuned task.
    \item Alternatively we can target the embedding of a finetuned transformer model. In this case the MockingBERT instance is no longer task agnostic, but can potentially better match the embeddings expected by the transformer model.
\end{itemize}
In our experiments we evaluate both of these approaches to understand the tradeoffs.

The models are trained on 64000 randomly selected sentences from BookCorpus for 5 epochs. The loss function is a combination of the mean squared error (MSE) loss for the subword boundary detection task and the MSE loss for the embedding prediction task. The loss values for the two components are scaled to have the same magnitude. This is to prevent one of the subtasks from dominating the other one during training.

During evaluation and inference, the input text is converted to lower case and the characters are embedded as a 768-dimensional vector representation.
Subsequently the MockingBERT model is executed to obtain subword embeddings for the input sequence. The embeddings are then bookended by the fixed representations of the transformer's special tokens, [CLS] and [SEP] in the case of BERT and <s> and </s> for RoBERTa. Finally, the sequences of embeddings are passed to the finetuned transformer model.

The transformer models are based on the pre-trained BERT Base and RoBERTa Base models and finetuned using the HuggingFace Transformers package \citep{wolf-etal-2020-transformers}.

For finetuning we attach a linear layer on top of the [CLS] (for BERT) or <s> (for RoBERTa) output embedding, and train the entire model using cross-entropy loss. We restrict the sequence length to 128 tokens, and the model is trained with a batch-size of 32, learning rate of 2e-05, for up to 5 epochs. We used an early stopping patience of 10, evaluated every 100 training steps.

\section{Proposed Attacks}

We propose \textsc{WordScoreAttack} , a cost effective way to generate adversarial attacks which can occur in an real-world setting. Typically trained models are hosted and exposed through an API and the user can only query the models by sending input and receiving output. Thus we only focus on black-box attacks, where one does not have access to the trained model. 

\textsc{WordScoreAttack} intelligently selects input tokens to perturb in order to maximize the chances of finding an adversarial example with the minimum number of perturbations. This is achieved by computing per-word conditional class probabilities. For a binary classification task we compute the log likelihood of each word as shown in equations 1 to 3, where $freq_{pos}, freq_{neg}$ are the frequencies of the word in positive and negative classes respectively. $N_{pos}, N_{neg}$ are the total words in the positive and negative corpus and $V$ is the total vocabulary size. We remove stop words and low frequency words before computing word scores. In the case of multi-class classification, for each class a separate score is computed by considering that class as positive and combining all other classes into the negative class.
\begin{equation} 
    word\_score=log(\frac{P(word_{pos})}{P(word_{neg})}) \\
\end{equation}
\begin{equation} 
    P(word_{pos}) = \frac{freq_{pos}+1}{N_{pos} + V}
\end{equation}
\begin{equation} 
    P(word_{neg}) = \frac{freq_{neg}+1}{N_{neg} + V}
\end{equation}
 
Given an input text and the original predicted class, \textsc{WordScoreAttack} targets words in the input text which have the highest scores for the given class and perturbs them in the decreasing order of scores, until the model prediction is flipped. Our hypothesis here is that the words with the highest scores are critical to the model's predictions and perturbing them can fool the model to make the wrong classification. 

We impose a query budget setting expressed by two parameters: \textit{(max\_tokens\_to\_perturb,  max\_tries\_per\_token)}. 
The first one denotes the maximum number of tokens that can be perturbed for each input text. The second parameter denotes the number of perturbation attempts allowed per token. When the \textit{max\_tries\_per\_token} for a token are exhausted and do not yield a successful attack, the attack greedily preserves the perturbation that decreases the model confidence the most (i.e. the one with the lowest score for the original predicted class) and moves to the next token in the order of word scores. The maximum number of model queries per input text are  \textit{max\_tokens\_to\_perturb * max\_tries\_per\_token}.

In order to mimic real world misspellings, we only allow one perturbation per input word. The perturbations considered are adding, deleting, and swapping of characters, as well as merging (deleting whitespace) and splitting of words (adding whitespace). Furthermore for the non-whitespace perturbations only the internal letters of a token are perturbed, and the first and last letters remain unmodified. This ensures the perturbed text can be comprehended by humans \citep{rawlinson1976significance,pruthi-etal-2019-combating}.

Though we allow one perturbation per token, multiple tokens per input text can be perturbed until a successful attack is found.
This is in contrast to \cite{pruthi-etal-2019-combating}, where the authors do an exhaustive search to find a single character perturbation that flips the model prediction.

\textsc{WordScoreAttack} finds an adversarial perturbation with a significantly lower query budget compared to an exhaustive search. This makes it a practical and effective way for both simulating adversarial attacks to analyse model resilience as well as for constructing adversarial samples for data augmentation and adversarial training.


\section{Experiments}

\begin{table*}
\small
\centering
\begin{tabular}{lccccc}
\toprule
\textbf{Model} & \textbf{IMDb} & \textbf{SST-2} & \textbf{SST-5} & \textbf{Yelp-2} & \textbf{Yelp-5} \\
\hline
\multicolumn{6}{l}{\textbf{BERT}} \\
(no remediations) & \textbf{88.0}/60.6/58.6 & \textbf{91.4}/42.2/38.8 & \textbf{56.2}/5.8/5.0 & 95.2/71.2/70.4 & \textbf{61.8}/27.2/25.0 \\
with RobEn \textsc{ConnComp} & 77.6/69.2/52.6 & 69.4/64.6/27.4 & 33.4/28.2/7.6 & 86.6/80.6/64.8 & 44.8/37.4/23.8 \\
with RobEn \textsc{AggClust} & 78.6/\textbf{72.0}/53.8 & 75.8/\textbf{72.6}/33.8 & 41.0/\textbf{34.6}/6.0 & 90.4/86.4/71.0 & 52.4/\textbf{42.6}/26.0 \\
\hline
\multicolumn{6}{l}{\textbf{\vtop{\hbox{\strut MockingBERT with BERT }}}} \\
targeting generic embedding & 86.8/70.6/\textbf{69.0} & 86.2/57.0/\textbf{56.8} & 51.6/10.8/\textbf{13.2} & 95.2/\textbf{88.6}/\textbf{89.8} & 60.0/40.8/\textbf{41.0} \\
targeting finetuned embedding & 86.4/69.8/68.6 & 86.8/57.4/\textbf{56.8} & 49.0/10.8/9.4 & \textbf{95.4}/88.4/89.0 & 61.2/40.2/40.2 \\
\hline
\multicolumn{6}{l}{\textbf{RoBERTa}} \\
(no remediations) & \textbf{90.8}/68.4/69.6 & \textbf{93.2}/48.6/46.8 & \textbf{57.2}/8.2/7.8 & 96.4/76.0/78.0 & \textbf{63.8}/36.8/36.0 \\
with RobEn \textsc{ConnComp} & 68.4/62.6/56.0 & 75.0/70.0/37.6 & 33.6/30.4/5.8 & 88.0/80.6/68.6 & 50.0/41.6/31.4 \\
with RobEn \textsc{AggClust} & 75.8/71.6/60.8 & 78.2/\textbf{74.8}/40.6 & 39.8/\textbf{34.8}/7.6 & 91.4/87.2/75.2 & 56.8/\textbf{48.8}/36.6 \\
\hline
\multicolumn{6}{l}{\textbf{\vtop{\hbox{\strut MockingBERT with RoBERTa }}}} \\
targeting generic embedding & 87.2/75.4/\textbf{77.0} & 88.8/62.0/\textbf{62.2} & 52.4/16.2/15.0 & 96.0/\textbf{88.0}/\textbf{88.2} & 62.8/47.0/\textbf{47.4} \\
targeting finetuned embedding & 87.8/\textbf{76.2}/76.0 & 90.0/60.4/\textbf{62.2} & 52.8/18.6/\textbf{17.0} & \textbf{96.6}/87.4/\textbf{88.2} & 62.6/46.8/45.8 \\
\bottomrule
\end{tabular}
\caption{\label{results_accuracy}
For each combination of model and task we provide an accuracy score for the following three variations of the test set: An unperturbed test set; a test set using \textsc{WordScoreAttack} excluding whitespace modifications; and a test set using \textsc{WordScoreAttack} including whitespace modifications. The reason for providing a score when excluding whitespace modifications is that some of the remediations have not been designed to counteract whitespace perturbations.
}
\end{table*}


We evaluated each model's performance on an unperturbed version of the test sets, as well as on \textsc{WordScoreAttack}. We also evaluated the finetuned BERT model on the IMDb task using the exhaustive adversarial attack as described in \cite{pruthi-etal-2019-combating}. Due to the prohibitive computational cost of this attack we were not able to evaluate it on the other models.

We evaluated variations both including and excluding whitespace modifications. This is because the remediations proposed in \cite{jones-etal-2020-robust} have not been specifically designed for combating such modifications.

For \textsc{WordScoreAttack}, we allowed the attack to change up to ten words, with edit distance of one. For each word, up to four attempts were made in order to find the one that decreased the model's confidence the most. As before, we evaluated both variants that allow and disallow whitespace modifications.

To establish baselines, we evaluated finetuned BERT and RoBERTa models with their default tokenizers, both with and without data augmentation. We also evaluated the \textsc{ConnComp} and \textsc{AggClust} approaches proposed in \cite{jones-etal-2020-robust}. Accuracy is used as the primary evaluation metric.

\begin{figure}[ht]
    \centering
    \makebox[\columnwidth]{\includegraphics[width=7.7cm]{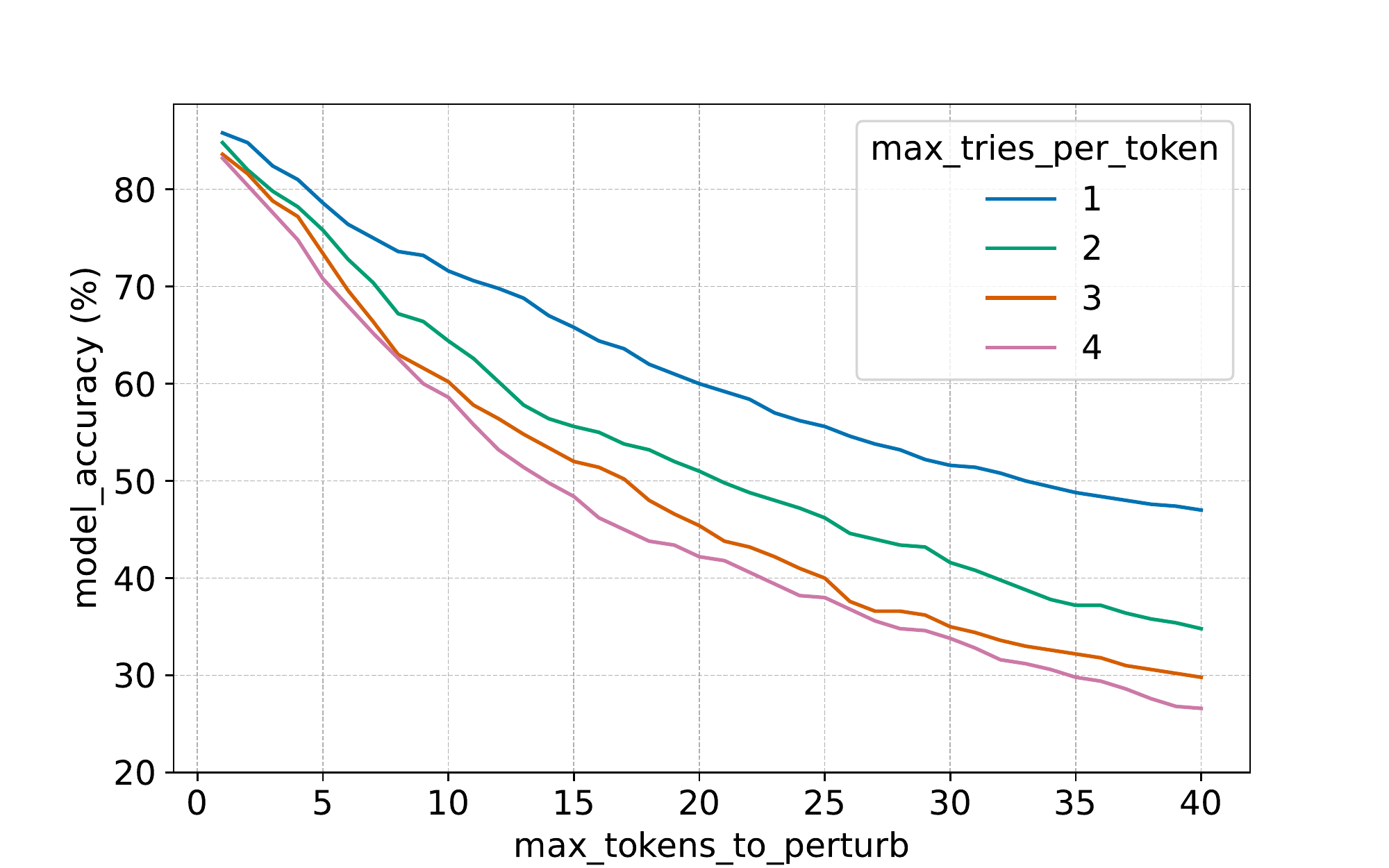}}
    \caption{\textsc{WordScoreAttack} analysis with different budget parameter settings.}  
    \label{fig:word_score_attack}
\end{figure}

A second stream of experiments is focused on evaluating the efficacy and efficiency of \textsc{WordScoreAttack}. We attack the BERT model finetuned for the IMDb task and vary the \textit{max\_tokens\_to\_perturb} from 1 to 40 and \textit{max\_tries\_per\_token} from 1 to 4. For each setting, we calculate the model accuracy on the 500 reviews in the test set. The results are shown in Figure \ref{fig:word_score_attack}. The original accuracy on the test set is 88\%. which is reduced to 26.6\% in the most adversarial setting of (40, 4). 

In order to compare \textsc{WordScoreAttack} to the exhaustive adversarial attack of \cite{pruthi-etal-2019-combating}, we evaluate a forgetful mode for \textsc{WordScoreAttack}, where an unsuccessful perturbation is reset when the attack moves to a new token. In this mode, the attack tries to flip the model's prediction by making only one perturbation to the input text. The results are shown in Table \ref{tab:wordscore_vs_pruthi}, where the normal mode is denoted as WSA and the forgetful model is denoted as WSA-Forgetful. Both modes operate with the query budget setting of (40,4). As expected the efficacy of the attack suffers in forgetful mode, with only a 15\% attack success rate and model accuracy dropping to only 72.8\% from the original 88\%. The exhaustive adversarial attack \citep{pruthi-etal-2019-combating} on the other hand reduces the accuracy to 66.8\%.  However this comes at the expense of 1,574 queries on average per attack compared to only 133 queries for WSA-Forgetful. In comparison the normal mode of WSA reduces the model accuracy to 26.6\%, with an attack success rate of 70\% while only requiring 83 queries on average.   

\begin{table}
\small
\centering
\begin{tabular}{lccc}
\toprule \textbf{} &  \textbf{Pruthi} & \textbf{WSA-Forgetful} & \textbf{WSA}\\ \hline
Orig. Accuracy & 89.6\% & 88\% & 88\% \\
Attack Success \% & 25.45\% & 18\%  & 70\%\\
Final Accuracy & 66.8\% & 72.6\% & 26.6\%\\
Avg. Queries & 1574 & 133 & 83\\
\bottomrule
\end{tabular}
\caption{\label{tab:wordscore_vs_pruthi} Comparison of exhaustive attack of Pruthi, 2019 with the forgetful (WSA-Forgetful) and normal  mode of \textsc{WordScoreAttack}(WSA). Pruthi and WSA- Forgetful are constrained to perturb only one character per input text. The budget setting for both WSA attacks is (40,4)}
\end{table}

\section{Analysis}

Our experiment results (Table \ref{results_accuracy}) show that MockingBERT consistently achieves the highest accuracy scores for adversarial attacks that allow whitespace modifications by a margin of between 5.6\% and 18.8\%. When whitespace modifications are disallowed, MockingBERT performs similarly to RobEn, with the exception of the SST datasets where the RobEn models achieve noticeably higher accuracy. It should be noted that pretraining MockingBERT on adversarial data with whitespace modifications results in noticeably lower performance on test sets without such modifications, as noted in the Ablation Studies section below.


Crucially, our model's accuracy on unperturbed data is typically only slightly lower than when using the standard BERT or RoBERTa tokenization/embedding procedure, with the accuracy scores ranging from being 0.2\% better to being 4.6\% worse than the standard model. This is in contrast with RobEn, where the accuracy on unperturbed data is lower by between 4.8\% and 17.4\%.


Interestingly the version of the tokenizer that targets default pre-trained transformer embeddings is performing slighly better overall than the version targeting embeddings finetuned specifically for individual tasks. This suggests that a universal tokenizer and embedder model can be used for a variety of tasks, with no need to adapt it specifically for each task. This hypothesis requires more research on a wider variety of tasks.



Through our analysis of the \textsc{WordScoreAttack}, we demonstrated that it is a cost effective way of constructing adversarial examples. It provides a flexible framework for evaluating models under attack. As shown in Figure \ref{fig:word_score_attack} model accuracy drops steadily as the \textit{max\_tokens\_to\_perturb} is increased. Accuracy declines faster with higher values of \textit{max\_tries\_per\_token} as more random perturbations can be tried per token.


The cost effectiveness of the attack is borne by the fact that the most adversarial budget setting of (40,4) requires an average of 83 model queries and reduces the accuracy to 26.6\%. On the other the setting of (20,1) will require at most 20 queries on average but still reduces the model accuracy to 60\%.



We wanted to construct adversarial samples such that humans can infer the original intent without much difficulty. Thus we considered a limited set of perturbations with constraints such as perturbing only one internal character per token, in addition to adding/deleting whitespace. Two examples of original inputs and their adversarial counterparts are given in Table \ref{tab:adversarial_examples} in the appendix.

\subsection{Ablation Studies}

To understand how data augmentation impacts the performance of MockingBERT, we have trained and evaluated variants of our approach trained on unperturbed training sets and on training sets where no whitespace perturbations were allowed. As with the main experiment we considered MockingBERT variants mimicking both the subword embeddings of generic (that is not finetuned) transformer models, as well as those mimicking finetuned transformer models. For comparison we have also finetuned a standard BERT model using data augmentation using the same perturbations as when training MockingBERT.

For experiments in this section we have used BERT as the underlying transformer architecture and we have evaluated it on the IMDb test set. The results (Table \ref{results_accuracy_ablation}) suggest that MockingBERT models require data augmentation to develop robustness to attacks. Indeed without data augmentation the accuracy of MockingBERT based models is strictly lower than that of pure BERT models. However MockingBERT based models trained with data augmentation significantly outperform similarly trained pure BERT models in the presence of test set perturbations.

Interestingly it appears that the presence of whitespace perturbations in the training set negatively affects the MockingBERT models' accuracy on test sets without such perturbations
(74.4\% vs. 70.6\% for the generic embedding variant and 72.8\% vs. 69.8\% for the finetuned embedding variant). Conversely it appears that including perturbed data when finetuning the pure BERT model improves the model's accuracy on unperturbed test data. A possible explanation is that in the presence of misspellings the transformer model learns to look for multiple redundant signals for its classification, in a way similar to the effects of dropout.

\begin{table}
\small
\centering
\begin{tabular}{lc}
\toprule
\textbf{Model} & \textbf{IMDb} \\
\hline
\multicolumn{2}{l}{\textbf{BERT}} \\
no augmentation & 88.0/60.6/58.6 \\
augmentation (incl. w/s) & \textbf{88.4}/53.6/55.4 \\
\hline
\multicolumn{2}{l}{\textbf{MockingBERT targeting generic embedding}} \\
no augmentation & 86.4/56.4/53.6  \\
augmentation (no w/s) & 87.2/\textbf{74.4}/60.2 \\
augmentation (incl. w/s) & 86.8/70.6/\textbf{69.0} \\
\hline
\multicolumn{2}{l}{\textbf{MockingBERT targeting finetuned embedding}} \\
no augmentation & 86.4/57.6/55.6  \\
augmentation (no w/s) & 86.4/72.8/58.8 \\
augmentation (incl. w/s) & 86.4/69.8/68.6 \\
\bottomrule
\end{tabular}
\caption{\label{results_accuracy_ablation}
Accuracy scores for models with various data augmentation strategies. The format is the same as in Table \ref{results_accuracy}.
}
\end{table}

\section{Future Work}


In our opinion it would be interesting to see if an approach similar to MockingBERT would work with other practical transformer-based models such as T5, XLNet or ELECTRA. Due to differences in how their tokenizers work some adaptations might be necessary. Nevertheless we see no fundamental issue that would prevent this approach from being applicable for these other model architectures.

Another potentially interesting direction would be to evaluate whether character-based transformer models such as Canine \citep{clark2022canine}, CharFormer \citep{tay2021charformer} or ByT5 \citep{xue2022byt5} are better suited to data augmentation with character-level perturbations than subword based transformer models. Intuitively subword based models are not well equipped for handling misspellings due to the fact that misspelled words might end up being mapped to unrelated tokens and that the number of possible misspellings for each token is very large. It should be noted that MockingBERT has a very clean separation between the tokenization/embedding procedure (which provides resilience to misspellings) and the main transformer-based language understanding layers. This means that it is easy to swap out the tokenizer and embedder if a new attack is devised, which may not be trivial for purely character-based transformer models.


\section{Conclusion}
We have demonstrated that our proposed MockingBERT embedder is able to successfully mimic the operation of a traditional tokenizer and embedder as used in the BERT and RoBERTa models, with only a modest decrease in performance on classification tasks. In the presence of input perturbations, MockingBERT outperforms both a data augmented BERT model and the state of the art RobEn procedure. Furthermore we have provided evidence that a universal embedder can achieve similar results to one that is specifically trained for a particular finetuned embedding, suggesting that embeddings might not need to be trained with specific tasks in mind. We have also proposed an efficient and effective method for constructing adversarial attacks, \textsc{WordScoreAttack}, which allows constructing such attacks at a fraction of the cost of an exhaustive search, at the expense of possibly perturbing more words within a sentence in order to achieve a similar attack success rate.

\section*{Acknowledgements}


The authors would like to thank prof. Christopher Potts and the course facilitators and staff of XCS224U for their support and feedback. 

Akash Singh would like to thank Salesforce Inc. for funding his participation in XCS224U.


We would like to thank the authors of the HuggingFace and TextAttack packages for making these highly useful tools freely available.


\section*{Authorship Statement}
The ideas presented in this article have been jointly developed and refined by all the authors. Akash Singh implemented \textsc{WordScoreAttack}, and performed the finetuning of the BERT model as well as the exhaustive adversarial attacks. Jan Jezabek implemented and trained the joint tokenizer and embedder models, and implemented the perturbers and the evaluation notebook.


\section*{Impact Statement}

Our main motivation for this research is to counteract intentional adversarial techniques designed to evade spam or toxic/obnoxious speech detection systems. We think that making such systems more robust results in more efficient moderation systems and more civil online discourse.

That said techniques such as MockingBERT can potentially be abused for surveillance and censorship purposes, by making it harder to fool and evade systems used for monitoring.

Similarly, \textsc{WordScoreAttack} can make it practical to evade text classification systems, regardless of whether such systems' goals can be considered noble or nefarious.

Training of the MockingBERT models described in this article took slightly over 20 hours of compute time on Nvidia Tesla K80 and Nvidia GeForce GTX 1070Ti GPUs. Evaluation of the models on \textsc{WordScoreAttack} took 25 hours, with an additional 50 hours of GPU time spent during development.

\bibliography{anthology,custom}
\bibliographystyle{acl_natbib}

\appendix
\section{Supplemental Data}
\label{sec:appendix}

\begin{table*}[t]\centering
\small
\begin{tabular}{|p{7.5cm}|p{7.5cm}|}
\hline \textbf{Original Input} & \textbf{Adversarial Input}   \\ \hline
the hand of death most definitely rates a ten on a scale of one to- due, in no small part, to john woo's masterful direction, coupled with kat's superb cinematography: some of the leisurely tracking shots alone are worth the price of a rental; there are moments when this one borders on becoming an art-house film. both james tien and sammo hung make for the kind of villains you can't help but love to hate. tien is particularly good as the baddest of the bad. it's a role reversal the likes of which i don't think i’ve ever seen before (tien normally played a hero and, in fact, with his, & the hand of death most defini tely rates a ten on a scale of one to- due, in no small part, to john woo's mastreful direction, coupled with kat's supreb cinematography: some of the leisurely tracking shots alone are worth the price of a rental; there are moments when this one borders on becoming an art-house film. both jtames tien and sammo hung make for the kind of villains you can't help but loe to hate. tien is particularly good as the baddest of the bad. it's a rloe reversal the likes of which i don't think i've ever seen before (tien normally played a hero and, in fact, with his  \\
 & \\
i caught this movie right in my eye when i was passing by a hall of posters in the nearby cinema. the tag line was sort of confusing and immediately after reading it, i thought of the possibility of it being similar to national lampoon's dorm daze. i liked that movie, aside from having a huge collection of such genres, i decided to hit it to the cinemas right after my exams for a tension releaser.<br /><br />delightfully, i came out smiling from cheek to cheek and had an equally great amount of laughter at bits and points of the movie. amanda aynes definitely kicked it off better than keira & i caughtthis movie right in my eye when i was passing by a hlal of posters in the nearby cin ema. the tag line was sort of confusing and immediately after reading it, i thought of the possibility of it being similar to national lampoon's dorm daze. i lciked that movie, aside from having a huge collection of such genres, i decided to hit it to the cinemas right after my exams for a tension releaser.<br /><br />delig htfully, i came out smliing from cheek to cheek and had an equallygreat amount of laughter at bits and points of the movie. amanda bynes definitely kicked it off better than keira \\

\hline
\end{tabular}
\caption{\label{tab:adversarial_examples} Examples of original inputs with adversarial counterparts.}
\end{table*}

\end{document}